\documentclass[conference]{IEEEtran}
\IEEEoverridecommandlockouts
\usepackage{cite}
\usepackage{amsmath,amssymb,amsfonts}
\usepackage{algorithmic}
\usepackage{graphicx}
\usepackage{textcomp}
\usepackage{xcolor}
\usepackage{algorithm}
\usepackage{float}
\usepackage{tabularx}
\usepackage{array}
\usepackage[table]{xcolor}
\usepackage{url}
\usepackage{gensymb}

\def\BibTeX{{\rm B\kern-.05em{\sc i\kern-.025em b}\kern-.08em
    T\kern-.1667em\lower.7ex\hbox{E}\kern-.125emX}}
\renewcommand{\IEEEauthorrefmark}[1]{\textsuperscript{#1}}
\begin{document}
\title{
A Framework for Automated Multi-Source Satellite Data Analytics and LLM-Based Report Generation
}

\author{
	\IEEEauthorblockN{Hind Yousif Alhammadi\IEEEauthorrefmark{1}, Isam Mashhour Al Jawarneh\IEEEauthorrefmark{2}}
\IEEEauthorblockA{\IEEEauthorrefmark{1}\textit{Department of Applied Physics and Astronomy}, 
\textit{University of Sharjah},  
Sharjah, UAE \\
Email: U25108456@sharjah.ac.ae}
\IEEEauthorblockA{\IEEEauthorrefmark{2}\textit{Department of Computer Science}, 
	\textit{University of Sharjah}, P.O.Box. 27272 Sharjah, UAE \\
	Email: ijawarneh@sharjah.ac.ae}
}

\makeatletter
\def\ps@IEEEtitlepagestyle{%
  \def\@oddfoot{\mycopyrightnotice}%
  \def\@evenfoot{}%
}
\def\mycopyrightnotice{%
  {\footnotesize
  \begin{minipage}{\textwidth}
  \centering
  © 2026 IEEE. Personal use of this material is permitted.  Permission from IEEE must be obtained for all other uses, in any current or future media, including reprinting/republishing this material for advertising or promotional purposes, creating new collective works, for resale or redistribution to servers or lists, or reuse of any copyrighted component of this work in other works.
  \end{minipage}\hfill}%
  \gdef\mycopyrightnotice{}
}
\makeatother

\maketitle
\bstctlcite{IEEEcontrol}
\begin{abstract}
This paper presents the workflow for building an automated ArcGIS Pro tool using ArcPy to extract the Land Surface Temperature (LST) from Landsat 7, 8 and 9 datasets. The tool eliminates the need for manual band selection and repetitive raster computations by automating the multi-step workflow of radiometric calibration, NDVI-based emissivity correction, and thermal conversion. In addition to supporting batch and single-scene processing, the tool has an optional Large Language Model (LLM) for statistical result interpretation and reporting. Depending on batch size, the tool reduced the processing time from around 11–58 minutes when done manually to around 4–11 minutes using the tool. We tested the tool with data from Ras Al Khaimah (RAK) in the UAE, and the LST obtained for Ras Al Khaimah ranged from approximately 25\textdegree C to 50\textdegree C, demonstrating an accurate LST mapping compatible with the weather conditions of RAK. In summary, our tool reduces human errors and improves processing accuracy and efficiency for thermal and environmental remote sensing applications, in addition to providing an interactive LLM-based interface for result's interpretation.
\end{abstract}


\section{Introduction}
Land Surface Temperature (LST) is one of the most significant factors in investigating climate change, urban climate analysis, vegetation assessment, and a wide range of environmental studies. LST is a measurement of Earth’s surface temperature. It is calculated from satellite imagery and used in a variety of scientific fields, including agriculture, hydrology, ecology, and meteorology. LST can help in identifying environmental issues, evaluate the severity of droughts, and measure the intensity of heat by using various analytical methods \cite{ almouctar2024drought,hu2023spatial}. Several studies on thermal analysis have been conducted using MODIS, ASETR, Landsat Thematic Mapper (TM)/Enhanced Thematic Mapper (ETM), and Landsat 8 data \cite{kowe2025spatial,patel2024urban}.

Typically, a raster calculator is used to obtain LST from a Landsat imagery in ArcGIS Pro. It performs a sequence of corrections to the thermal band, that mainly consists of the following: conversion of Top of Atmosphere (TOA) Radiance, conversion to Top of Atmosphere (TOA) brightness temperatures, and conversion from At-Satellite Temperature to Land Surface Temperature. These steps usually require repeated use of the raster calculator, which increases the risk of user error in addition to increasing analysis processing time. Few satellites provide calibrated LST, such as Sentinel 3 and MODIS, however, they have lower resolutions (capped at approximately 1km). This low resolution is typically insufficient for research that requires more precise spatial information in the form of a higher resolution. Consequently, developers and researchers frequently rely on a higher resolution Landsat's 30m thermal data, which typically imposes a higher processing time.

This project addresses this problem by developing an automated LST tool in ArcGIS Pro using ArcPy. The tool accepts Landsat 7, 8, and 9 data files and generate the LST map directly, without running the raster calculator or even opening the data files to extract the required band. The tool can recognize each satellite from the file name and uses the specific constants required for the calculations. We have also integrated a Large Language Models (LLM)-based interface to produce a summarized report based on LST map's statistics. Our LST tool improves productivity, consistency, and utility for academics and researchers using thermal remote sensing data by automating the entire workflow.
The specific objectives of this study are the following: (1) To automate the process of calculating LST and supporting Landsat 7, 8, and 9. (2) to allow single-scene and batch processing. (3) to support LLM-driven reporting and interpretation. (4) To reduce human error in calculating LST and improve the processing accuracy and efficiency.

\section{Literature Review}
Land Surface Temperature (LST) has gained widespread recognition as an essential factor in climate and environmental research due to its influence on surface energy exchange and climatic variability.  To generate LST values, recent studies utilize Landsat 8 and 9 satellite data. They employ Single-Channel methods on Band 10 or multiple channel Split-Window (SW) algorithms that utilize both Thermal Infrared Sensor (TIRS) Bands 10 and 11 \cite{11422992, ASGHARISARASKANROUD20242915, Nugraha31122024}. The algorithms studied include methods based on the Radiative Transfer Equation (RTE), particularly Single-Channel techniques, and Multi-Channel Split-Window Algorithms (SWA) \cite{11422992, ASGHARISARASKANROUD20242915, Nugraha31122024}. Recent studies have shown techniques based on NDVI-derived emissivity values, where they change raw digital numbers to radiance, reflectance, and brightness temperature before synthesizing LST using the Radiative Transfer Equation (RTE) approach with Landsat 8 data \cite{Nugraha31122024, ASGHARISARASKANROUD20242915}. Similarly, recent studies have employed the Split-Window approach using Landsat 8 thermal bands, and they demonstrate that LST estimates differ depending on emissivity and atmospheric correction \cite{Nugraha31122024, ASGHARISARASKANROUD20242915}.

 Numerous studies have used NDVI-based emissivity models to address emissivity, which remains a major concern in LST extraction. Sekertekin et al. used Surface Radiation Budget Network (SURFRAD) ground measurements for validation to assess many LSE models and retrieval techniques in Landsat 5, 7 and 8 in rural locations \cite{Sekertekin2020}. The MWA approach produces the lowest error (RMSE 2.39 K) throughout missions, according to the study's conclusion that algorithm accuracy differs by sensor type. The study also demonstrates the growing interest in automated, tool-based processes by offering an ArcGIS toolbox that integrates many retrieval techniques.

The use of LST analysis in environmental monitoring is the subject of other research. \cite{Mohan2021} used Landsat-derived trends from 2016 to 2020 and NDVI–LST correlations to evaluate drought severity. The findings confirmed the value of LST as an indicator of agricultural stress, showing a strong correlation (up to 99\%) between NDVI and LST . Similarly, \cite{Aryalekshmi2020} used Landsat 8 thermal bands in conjunction with supplementary vegetation indicators to study LST change in agricultural zones. The results demonstrated a significant agreement (RMSE 0.94–1.26°C) between satellite-estimated LST and meteorological station readings.
The use of traditional processing procedures, such as determining necessary bands, collecting metadata information, and repeatedly performing raster computations, is a common theme in the literature. While a number of studies suggest automated or semi-automatic procedures, the majority are either unable to process batch processing or restricted to single satellite types, or need manual band selection. Furthermore, many procedures still use Level-1 products to adjust intermediate variables such as NDVI, radiance, and emissivity, or to preserve consistency across historical archives.

\section{Methodology}
In this section, we explain the workflow for constructing an automated tool to calculate the LST, including data preparation, correction factors, and the complete pipeline. 

\subsection{Study Area}
Ras Al Khaimah (RAK) in the UAE, is one of the seven Emirates of the United Arab Emirates (UAE), is the study area that we have chosen for thispaper. It is located at 55° 60 E and 25° 26 N. It borders the Sultanate of Oman and is located in the northernmost region of the United Arab Emirates. It is estimated that RAK has 400,000 residents and a total area of 2478 square kilometers in 2025. 

\subsection{Input Data}
 Level-1 Landsat images from Landsat 7, Landsat 8, or Landsat 9 can be used as an input for our tool. Each scene input to the tool must use the file name generated by Earth Explorer, and the file must contain the thermal, red, and NIR bands, as well as metadata. This produces the scene's LST. Table 1 summarizes the band numbers that are required for each satellite, where Landsat 7 uses bands 3, 4, and 6 while Landsat 8/9 uses bands 4, 5, and 10 for Red, NIR, and thermal bands, respectively. The Red and NIR bands are required to calculate NDVI (see \ref{eq:ndvi}), which is then used for the emissivity correction. The thermal band is used to determine the brightness temperature. The radiometric scaling parameters that is needed to transform raw Digital Number (DN) measurements into radiance, reflectance, and final land surface temperature are provided by the Metadata Text file (MTL) metadatam which accompanies satellite imagery, particularly Landsat data products. It contains essential metadata information that is needed for processing the raw data. DM is a raw pixel value stored by satellite sensors. They are uncalibrated digital values that represent the intensity of electromagnetic radiation that is recorded by the sensor for each spectral band. Since level-2 items already have processed surface temperature information, they are not utilized.
 A Summary of the Landsat spectral bands that we have used in this paper are summarized in table \ref{tab:bands}

\begin{table}[ht]
\caption{Summary of the Landsat spectral bands used in this study.}
\label{tab:bands}
\centering
\begin{tabularx}{\columnwidth}{|>{\centering\arraybackslash}X|
                                >{\centering\arraybackslash}X|
                                >{\centering\arraybackslash}X|
                                >{\centering\arraybackslash}X|}
\hline
\rowcolor[gray]{0.7}
\textbf{Satellite/Bands} & \textbf{Red} & \textbf{NIR} & \textbf{Thermal} \\
\hline
\cellcolor[gray]{0.7}\textbf{Landsat 7} & \textbf{3} (0.63--0.69 µm), 30 m & \textbf{4} (0.77--0.90 µm), 30 m & \textbf{6} (10.40--12.50 µm), 60 m \\
\hline
\cellcolor[gray]{0.7}\textbf{Landsat 8} & \textbf{4} (0.64--0.67 µm), 30 m & \textbf{5} (0.85--0.88 µm), 30 m & \textbf{10/11} (10.6--11.19 µm), 100 m \\
\hline
\cellcolor[gray]{0.7}\textbf{Landsat 9} & \textbf{4} (0.64--0.67 µm), 30 m & \textbf{5} (0.85--0.88 µm), 30 m & \textbf{10/11} (10.6--11.19 µm), 100 m \\
\hline
\end{tabularx}
\end{table}

\subsubsection{Conversion of TOA Radiance}
The first step in the process of calculating the LST is to convert the thermal band digital number (DN) into Top of Atmosphere (TOA) radiance using Equation~\ref{eq:radiance}.

\begin{equation}
L_{\lambda} = M_L \times Q_{cal} + A_L
\label{eq:radiance}
\end{equation}

Where:  
\begin{itemize}
    \item $L_{\lambda}$ = TOA spectral radiance (Watts/(m\textsuperscript{2}·sr·$\mu$m))
    \item $M_L$ = band–specific multiplicative rescaling factor (from metadata)
    \item $Q_{cal}$ = quantized and calibrated pixel value (DN)
    \item $A_L$ = band–specific additive rescaling factor (from metadata)
\end{itemize}

\subsubsection{Conversion to Top of Atmosphere Brightness Temperature}

The second step converts TOA radiance to brightness temperature using Equation~\ref{eq:bt}.

\begin{equation}
T_B = \frac{K_2}{\ln{\left(\frac{K_1}{L_{\lambda}} + 1\right)}} - 273.15
\label{eq:bt}
\end{equation}

Where:
\begin{itemize}
    \item $T_B$ = top-of-atmosphere brightness temperature (Kelvin)
    \item $L_{\lambda}$ = TOA spectral radiance
    \item $K_1, K_2$ = band–specific thermal conversion constants (from metadata)
\end{itemize}

\subsubsection{Conversion from At-Satellite Temperature to Land Surface Temperature}

The final step is to convert the brightness temperature to the Land Surface Temperature (LST) using Equation~\ref{eq:lst}.  
Before applying this equation, the emissivity must be calculated using Equations~\ref{eq:emissivity}–\ref{eq:ndvi}.
\begin{equation}
T = \frac{T_B}{1 + \left(\frac{\lambda T_B}{c_2}\right) \ln{\left(\varepsilon\right)}}
\label{eq:lst}
\end{equation}

Where:
\begin{itemize}
    \item $\lambda$ = wavelength of emitted radiance
    \item $c_2 = \frac{hc}{k} = 1.4388 \times 10^{-2} \, \text{m·K}$
    \item $h$ = Planck's constant ($6.626 \times 10^{-34}$ J·s)
    \item $k$ = Boltzmann constant ($1.38 \times 10^{-23}$ J/K)
    \item $c$ = speed of light ($2.998 \times 10^{8}$ m/s)
    \item $\varepsilon$ = surface emissivity
\end{itemize}

Surface emissivity is estimated using the NDVI-based method in equation \ref{eq:emissivity}:

\begin{equation}
\varepsilon = 0.004 \times P_v + 0.986
\label{eq:emissivity}
\end{equation}

Where $P_v$ (proportion of vegetation) is calculated as in \ref{eq:pv}:

\begin{equation}
P_v = \left( \frac{NDVI - NDVI_{min}}{NDVI_{max} - NDVI_{min}} \right)^2
\label{eq:pv}
\end{equation}

The normalized difference vegetation index (NDVI) is calculated from the near-infrared (NIR) and red (RED) bands using equation \ref{eq:ndvi}:

\begin{equation}
NDVI = \frac{NIR - RED}{NIR + RED}
\label{eq:ndvi}
\end{equation}

These steps must be repeated several times and depending on each satellite or scene, the MTL metadata file must be opened to extract the required values to fill the equations, such as the K1 and K2 constants. Table 2 shows the summarized required data to be extracted from the metadata file to produce the LST. 

\begin{figure}[ht]
    \centering
    \includegraphics[width=\columnwidth,height=0.65\textheight,keepaspectratio]{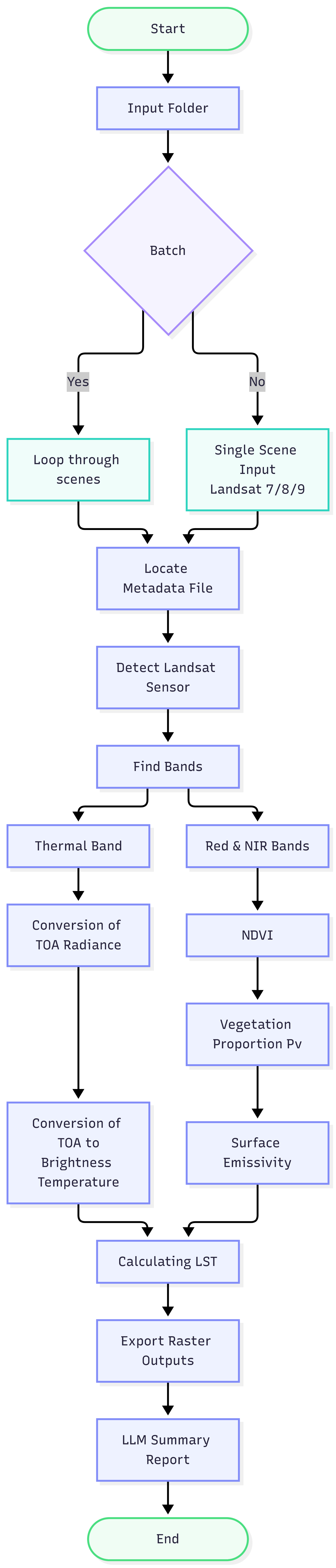}
    \caption{Flowchart showing the automated LST workflow.}
    \label{fig:flowchart}
\end{figure}

\begin{table} [ht]
\centering
\caption{Metadata field required to be used in equations.}
\label{tab:Metadata}
\begin{tabular}{| l | l |}
\hline
\rowcolor[gray]{0.7} \textbf{Metadata Feild} & \textbf{Use in equations} \\
\hline
RADIANCE\_MULT\_BAND\_X & ML in Equation 1 \\
\hline
RADIANCE\_ADD\_BAND\_X & AL in Equation 1 \\
\hline
K1\_CONSTANT\_BAND\_X & K1 in Equation 2 \\
\hline
K2\_CONSTANT\_BAND\_X & K2 in Equation 2 \\
\hline

\end{tabular}

\end{table}
The tool can also be used for batch processing (i.e., multiple files). We can prepare a parent file that contains multiple Landsat scenes to generate LST for each scene in the folder. If valid file extensions are found in the folder, they follow the workflow that we have discussed for single-scene processing, which is summarized in the flow chart shown in Figure \ref{fig:flowchart}. This can be useful for scenarios that require large-area thermal mapping. The time-consuming process of reviewing multiple scenes to obtain LST is automated, thus saving time and reducing errors.

The tool also includes an optional function that integrates a Large Language Model (LLM) to provide an interpretive summary of the output LST dataset. It collects basic descriptive statistics, including the minimum, maximum, mean, and standard deviation of LST values, along with raster metadata such as geographic reference, resolution, and scene name. This data is then submitted to the LLM via an API request. The generated summary provides a quick qualitative assessment of the scene's thermal characteristics, including potential heat islands, natural temperature gradients, and anomalous thermal patterns. This outcome is documented in a text-based report together with the raster products. If the user chooses to disable the capability or fails to provide an API key, the report is generated but contains only numerical data without interpretation. The LLM component enhances the usability of our system. It provides an automatically generated narrative that speeds up evaluation, decision-making, and the transmission of results to non-technical individuals. It does this without affecting the fundamental LST computation.

\section{System Design and Implementation}
\begin{figure}[H]
    \centering
    \includegraphics[width=\columnwidth, keepaspectratio]{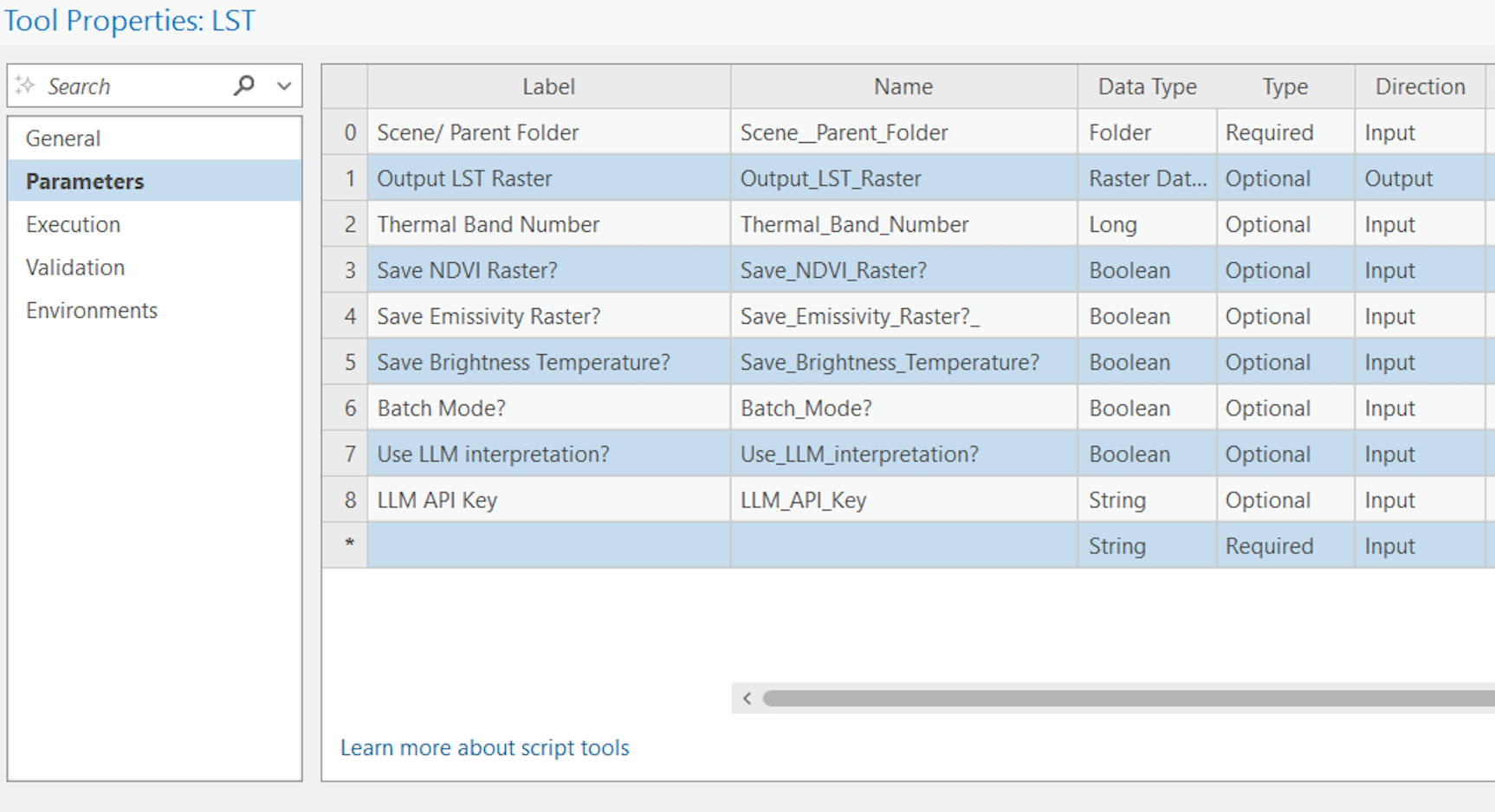}
    \caption{Parameters of LST Tool.}
    \label{fig:params}
\end{figure}

We have used ArcGIS Pro software to create the Land Surface Temperature (LST) tool by implementing a Python script through the ArcPy library. The tool contains 9 parameters, as shown in Figure \ref{fig:params}. The first parameter is where the user adds the input data files, and it can be a single Landsat scene or a parent folder containing multiple scenes. In the second parameter, the user can specify the output raster path. Additionally, the tool includes an optional thermal band selector that allows users to select a specific thermal band as needed. This applies to Landsat 8 and 9: choose between bands 10 and 11; otherwise, band 10 is automatically used. Followed by three parameters for optional saving of NDVI, emissivity, and brightness temperature rasters. The seventh parameter must be checked when adding batch scenes as input; otherwise, it is assumed that the input is a single scene. The last two parameters are for optional reporting using an LLM as follows: an API Key that must be obtained and provided to produce results. If left empty, a report file will be generated that contains LST results statistics without interpretation.  The tool integrates an LLM by sending an HTTP request via an API key, using statistical data (minimum, maximum, mean, and standard deviation) computed by ArcGIS Pro functions. The LLM uses these metrics to generate a short paragraph describing temperature patterns and any thermal extremes or irregularities, then Python scripts compile and save as a text-based report.

\section{Results and discussion}
\begin{figure}[ht]
    \centering
    \includegraphics[width=\columnwidth, keepaspectratio]{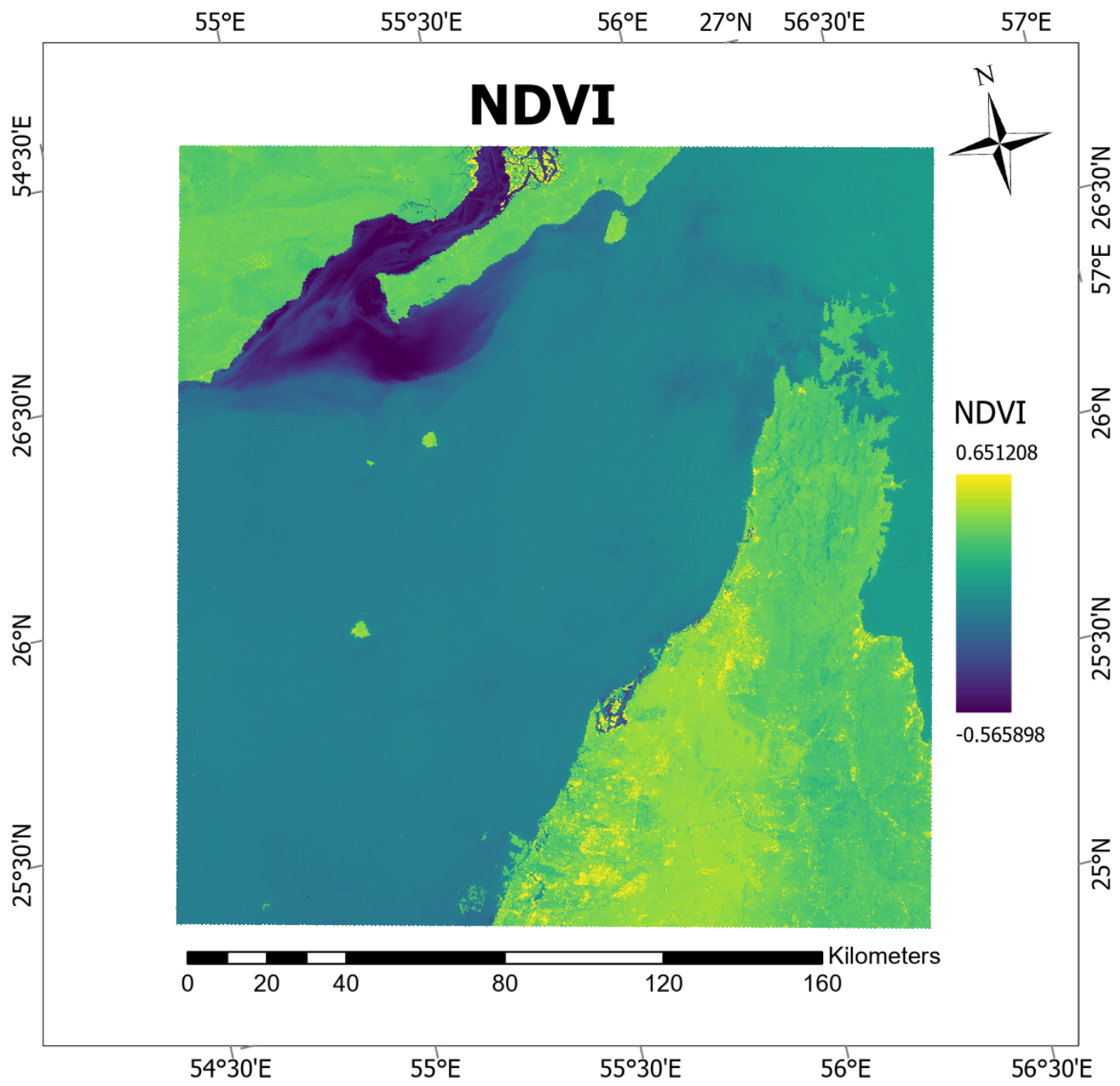}
    \caption{Obtained NDVI map for RAK using Landsat 8.}
    \label{fig:NDVI}
\end{figure}

\begin{figure}[ht]
    \centering
    \includegraphics[width=\columnwidth, keepaspectratio]{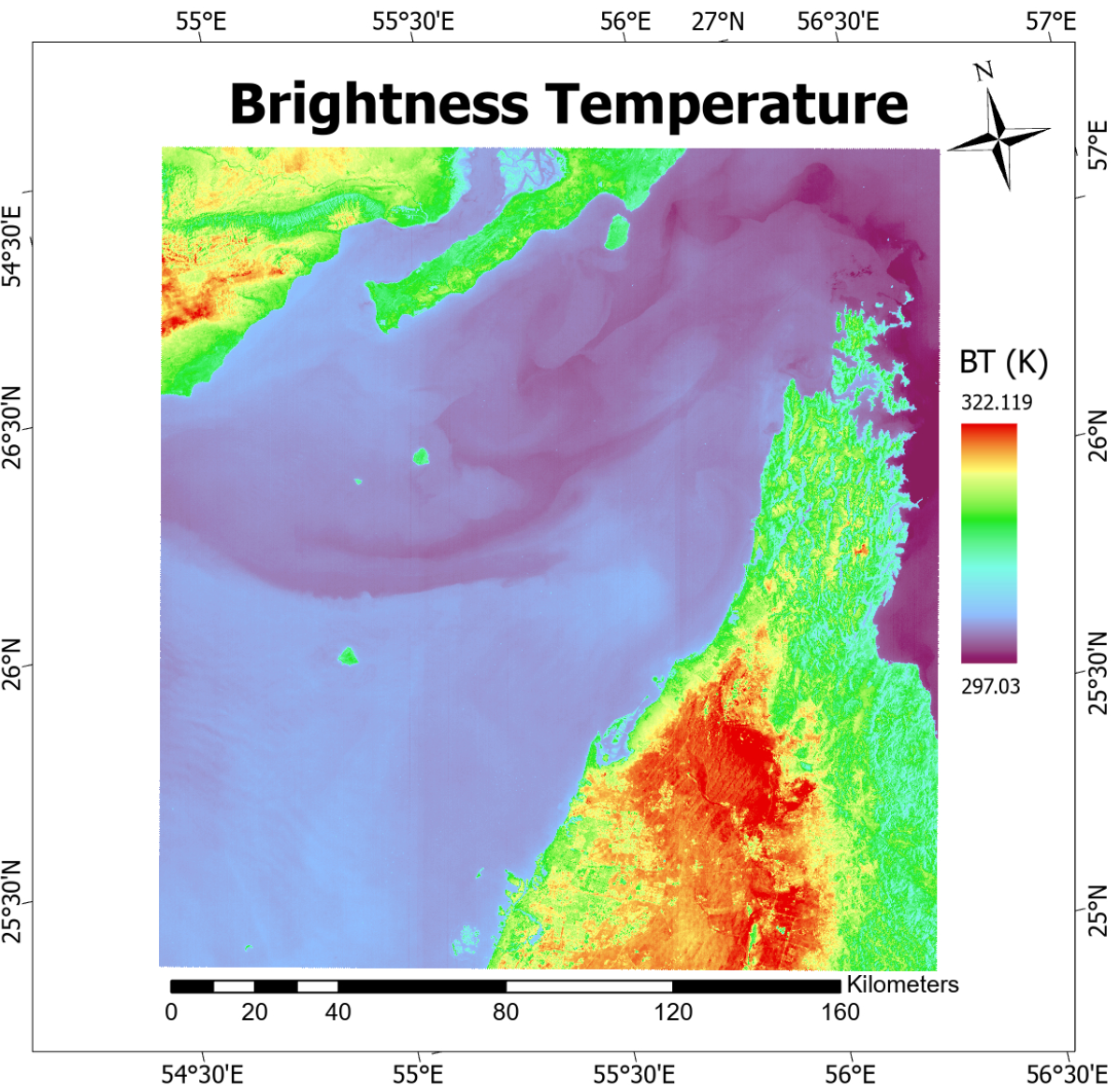}
    \caption{Obtained BT map for RAK using Landsat 8.}
    \label{fig:BT}
\end{figure}

\begin{figure}[ht]
    \centering
    \includegraphics[width=\columnwidth, keepaspectratio]{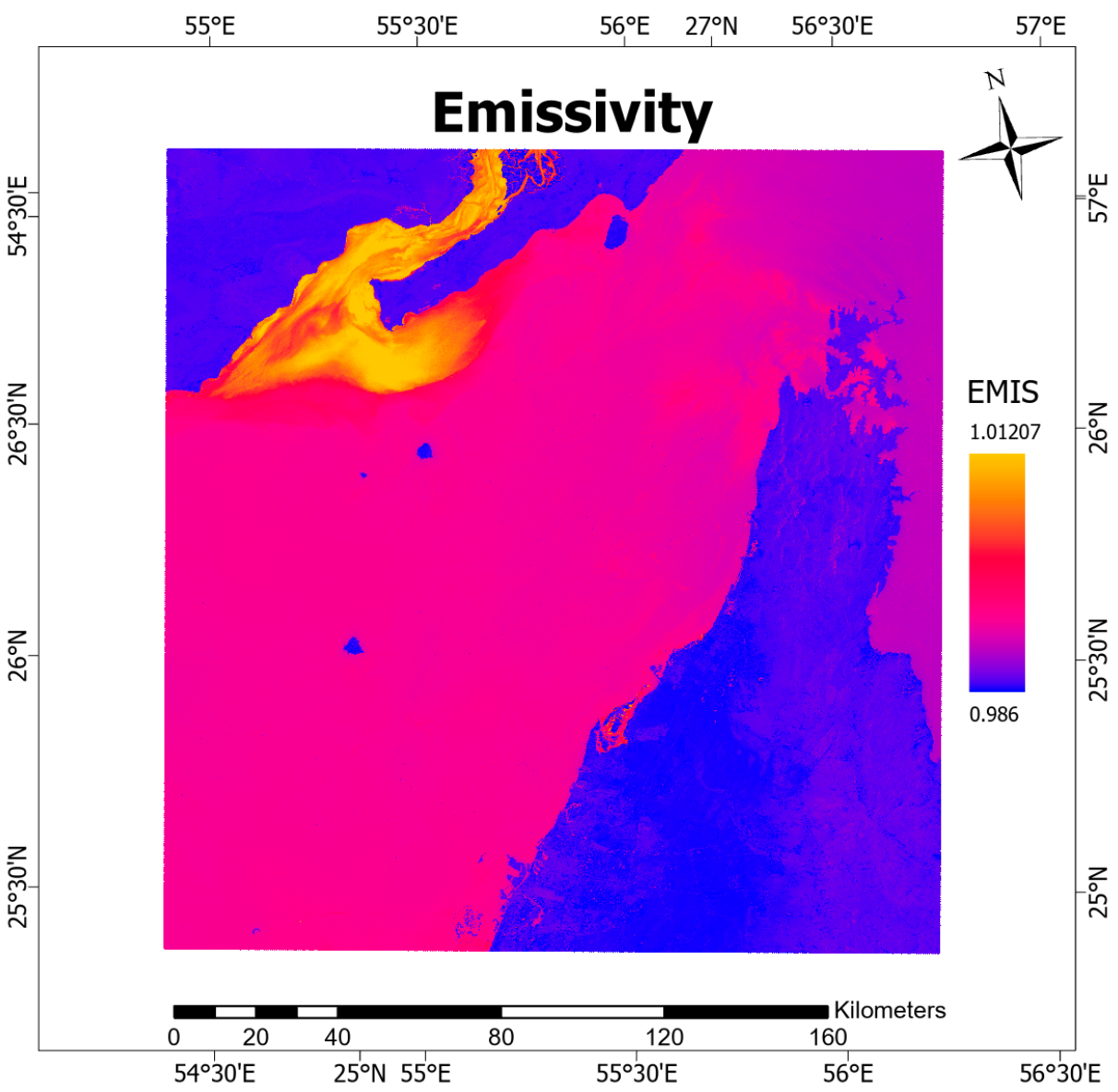}
    \caption{Obtained Emissivity map for RAK using Landsat 8.}
    \label{fig:Emissivity}
\end{figure}

\begin{figure}[ht]
    \centering
    \includegraphics[width=\columnwidth, keepaspectratio]{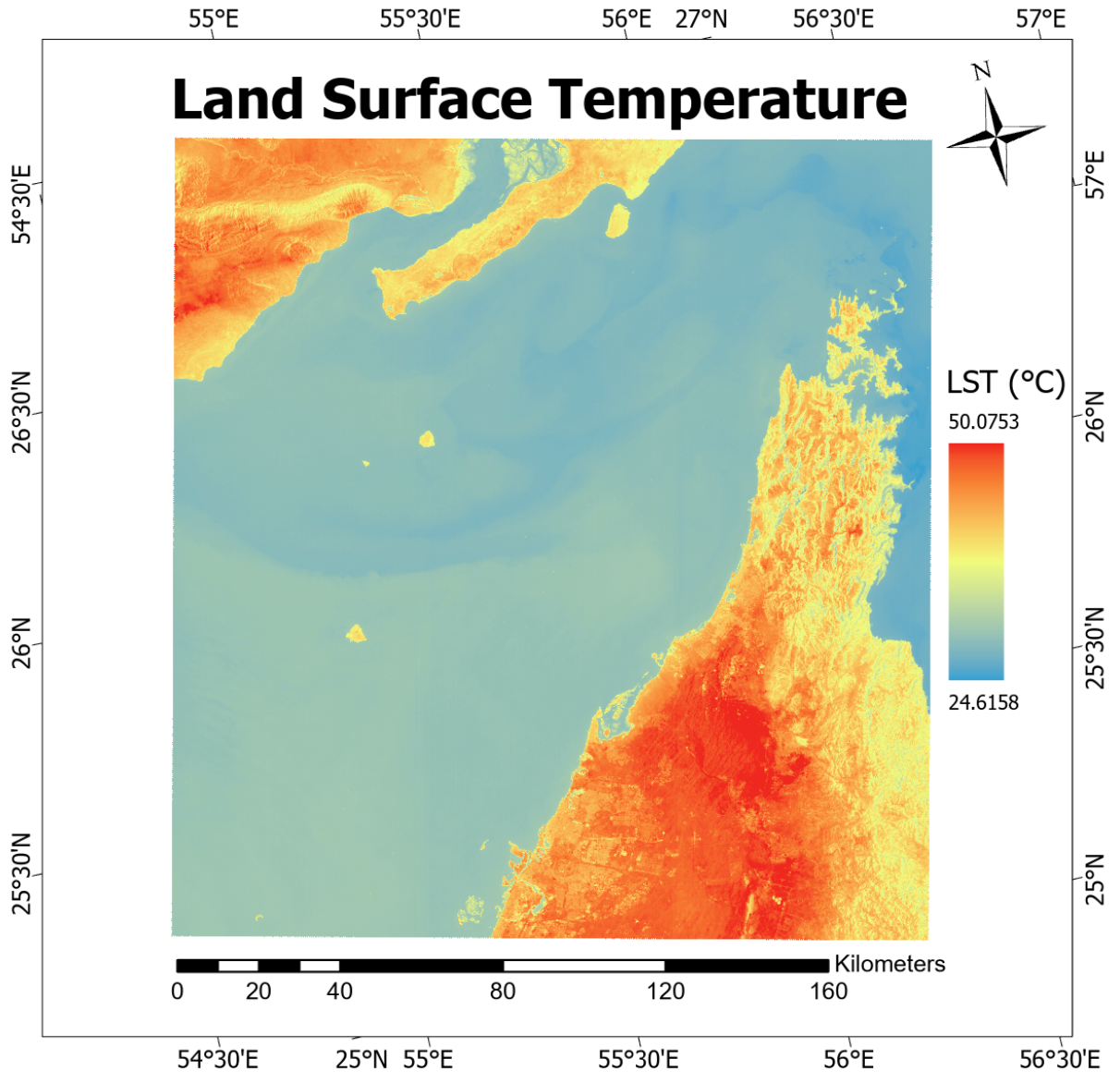}
    \caption{Obtained LST map for RAK using Landsat 8.}
    \label{fig:LST}
\end{figure}

Our tool was tested on all three satellites and we have verified that it works as planned. For demonstration purposes, the results obtained from Landsat 8 are presented in this paper. In addition to extracting the final Land Surface Temperature (LST), the developed tool automatically outputs and saves all intermediate rasters—namely NDVI, Brightness Temperature, and Surface Emissivity into the output directory, as these layers are sequentially required for the final LST calculation. Figures~\ref{fig:NDVI}, \ref{fig:BT}, and~\ref{fig:Emissivity} present these generated layers over the study area. Customized color ramps and symbology schemes were applied to each raster to optimize spatial interpretability. The LST map produced is shown in Figure \ref{fig:LST}. The data indicate that temperatures range from approximately 25 to 50 degrees Celsius, which is a reasonable range for the Emirates of RAK in the UAE. Also, the LLM-based report is shown in Figure \ref{fig:flowchart}, which includes raster information, LST Statistics, and an interpretation of the results.

\begin{figure}[ht]
    \centering
    \includegraphics[width=\columnwidth, keepaspectratio]{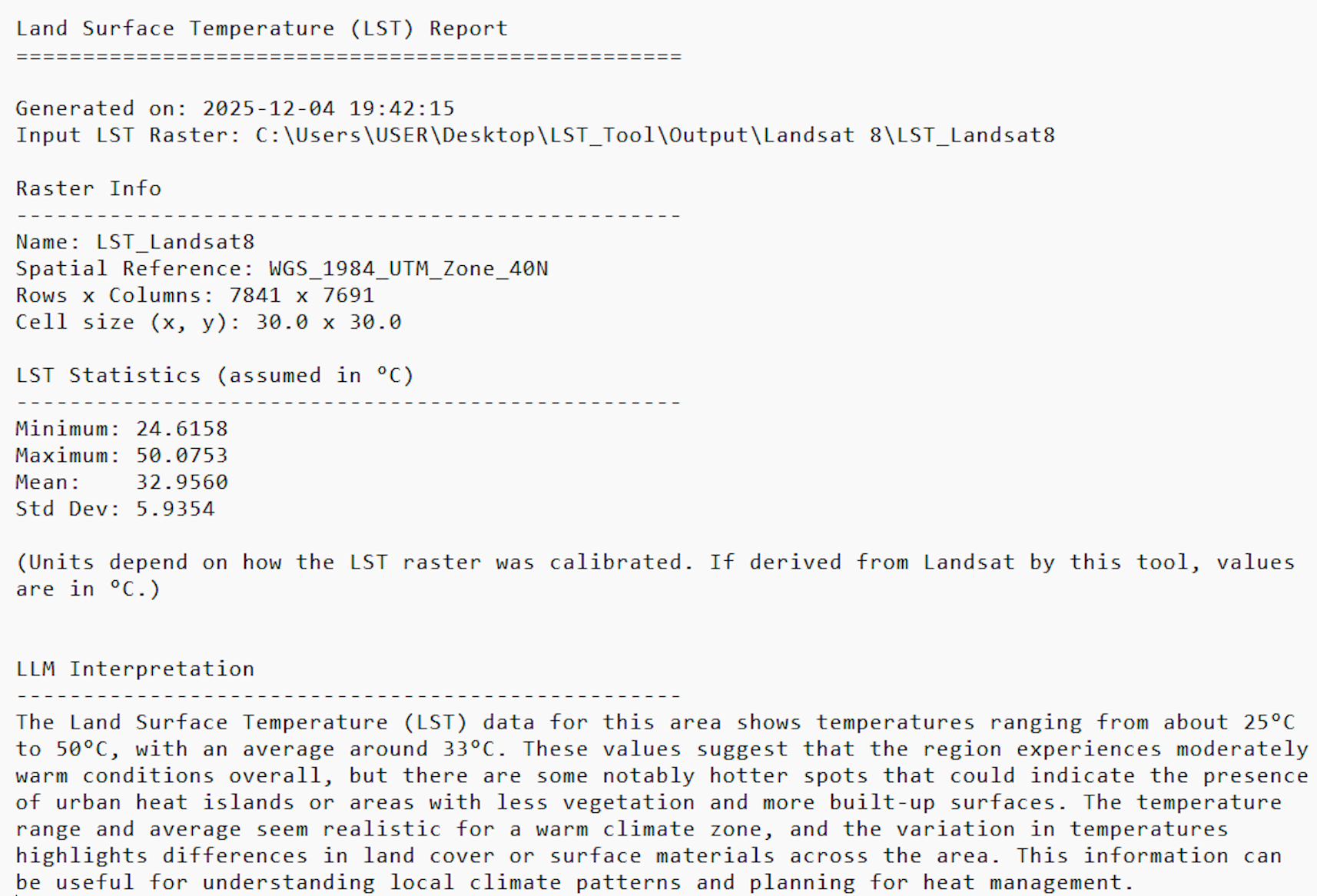}
    \caption{LST summary report generated using LLM.}
    \label{fig:report}
\end{figure}

To assess the tool's efficiency, its running time is compared with manual processing time, assuming the manual processing time remains constant and is multiplied by the number of scenes. The tool took 3m 52s to produce the output by running all optional parameters, including saving NDVI, emissivity, and brightness temperature, and using an LLM to generate a summary report of the data. Manually finding LST through the raster calculator can take up to 11m 33s or even longer if users are beginners or running a batch process. The time saved when finding LST for a single scene is 7m 41s, and the time saved grows with batch processing: running five scenes saves 46m 29s. The tool took 11m 16s to run a batch of scenes, which is less than the time required to manually find the LST for a single scene. Table 3 shows a comparison between the plain manual running time and our tool's running time for processing 1-5 scenes.

\begin{table}[ht]
\centering
\caption{Comparison between manual and automated LST processing time.}
\label{tab:performaL7 Scene 1 	12m 33s 	3m 41s	1.62 	0.88 nce}
\begin{tabular}{|l|c|c|c|c|}
\hline
\rowcolor[gray]{0.7} \# Scenes& Manual Time & Tool Time & Time Saved \\
\hline
1& 11m 33s& 3m 52s& 7m 41s \\
\hline
2& 23m 06s& 4m 31s& 18m 35s \\
\hline
3& 34m 39s& 7m 36s& 27m 03s \\
\hline
4& 46m 12s& 9m 3s& 37m 09s\\
\hline
5& 57m 45s& 11m 16s& 46m 29s\\
\hline
\end{tabular}
\end{table}

\section{Conclusion and Future Work}
In this study, we have built an automated LST extraction tool for ArcGIS Pro using the ArcPy. The tool supports Landsat 7, 8, and 9 datasets. Specifically, we have automated the workflow of manually identifying bands, extracting metadata values, and going through the raster calculator multiple times to finally extract the LST. Additionally, the tool supports single-scene and batch (i.e., multiple-scene) processing, thus reducing the processing time and minimizing the human error. Furthermore, the Large Language Model (LLM) LST-based report generation greatly enhances usability and increases the comprehensibility of the results for non-expert audiences. In a future work, we will integrate cloud masking techniques and atmospheric correction capabilities. Also, we will develop novel techniques that aim at reducing the running time, especially for batch processing, by integrating cloud-based computing or GPU acceleration. With further development, the tool can serve as a thermal remote sensing framework for environmental research, policy assessment, and GIS-based decision support systems. 

\bibliographystyle{IEEEtran}
\bibliography{export}

\end{document}